\documentclass{article} 
\usepackage{iclr2027_conference,times}

\usepackage{amsmath,amsfonts,bm}

\def\eqref#1{equation~\ref{#1}}

\def\1{\bm{1}}

\DeclareMathAlphabet{\mathsfit}{\encodingdefault}{\sfdefault}{m}{sl}
\SetMathAlphabet{\mathsfit}{bold}{\encodingdefault}{\sfdefault}{bx}{n}

\usepackage{hyperref}
\usepackage{url}

\usepackage{inconsolata}
\usepackage{graphicx}
\usepackage{multirow}
\usepackage{pifont}
\usepackage{siunitx}
\usepackage{amsmath}
\usepackage{amsthm}
\usepackage{xcolor}
\usepackage[dvipsnames,table]{xcolor}
\usepackage{colortbl}
\usepackage{hhline}
\usepackage{subcaption}
\usepackage{amsfonts}
\usepackage{svg}
\usepackage{graphicx}
\usepackage{bbm}
\usepackage{dsfont}
\usepackage{booktabs}
\usepackage{tabularx}
\usepackage{enumitem}
\usepackage{xltabular}
\usepackage[T1]{fontenc}
\usepackage{textcomp}
\usepackage{tcolorbox}
\tcbuselibrary{listings}
\lstdefinestyle{promptstyle}{
  basicstyle=\ttfamily\small, 
  breaklines=true,
  breakatwhitespace=false, 
  columns=flexible, 
  frame=single,
  xleftmargin=0pt, 
  framexleftmargin=0pt, 
  showstringspaces=false,
  upquote=true
}
\newcommand{\logo}[1]{\textbf{#1}}
\usepackage{enumitem}
\setlist{nosep, leftmargin=*} 

\title{Handle with CARE: Can LLMs Reproduce How Online Communities React?}

\author{
    Nuan Wen$^{\diamond\dagger}$, 
    Chanbin Lim$^\diamond$,
    Xuezhe Ma$^{\diamond\dagger}$, \\ 
    $^\diamond$Department of Computer Science, University of Southern California\\ 
    $^\dagger$Information Sciences Institute, University of Southern California\\ 
    \texttt{\{nuanwen, chanbinl, xuezhema\}@usc.edu}
}
\iclrfinalcopy 
\begin{document}

\maketitle

\begin{abstract}
Large language models (LLMs) are increasingly used as proxies for computational social analysis, yet faithfully representing the “thick descriptions” \citep{geertz1973interpretation} of human communities remains a critical challenge. Current evaluations often reduce social identity to static labels, sidelining how real-world groups navigate social shifts. To bridge this gap, we introduce \logo{CARE} (\logo{C}ommunity-\logo{A}ware \logo{R}eaction \logo{E}valuation), a reaction-centered framework that benchmarks LLM-simulated discourse against the authentic, event-contingent responses of distinct communities to real-world news. Spanning 207 Reddit communities and covering 9,947 authentic reactions towards 2,166 news articles, CARE evaluates leading LLMs using a hierarchical taxonomy covering coarse attitudes and fine-grained communicative tones. Our empirical findings expose two critical failure modes in prevailing community-conditioning paradigms. First, while community context and targeted reasoning significantly enhance macro-level attitudinal and tonal profiling, these gains largely collapse at the instance level when predicting reactions to specific events. Second, the benefits of community conditioning are remarkably uneven: prompting strategies yield non-uniform shifts, where fidelity gains in some communities are offset by performance drops in others. This micro-macro divergence and community-level instability demonstrate that standard conditioning enables models to approximate static baseline profiles without capturing dynamic or equitable event reactions, establishing CARE as an essential diagnostic tool for community-aware social simulation.\footnote{Code and data are available at \url{https://github.com/nuankw/Handle_with_CARE}.} \looseness=-1
\end{abstract} 
\section{Introduction}

Large language models (LLMs) are increasingly deployed as generative proxies for social analysis, celebrated for their ability to encode the linguistic contours of social norms and scale the simulation of human dynamics \citep{anthis2025position, piao2025agentsociety}. Yet, as their application in computational social science expands, a critical disconnect has emerged between how models are aligned to represent social groups and how these groups authentically function \citep{hovy-yang-2021-importance, kabir2025break, shen-etal-2025-revisiting, wang2025large}. Anthropological and sociolinguistic traditions emphasize that communities are not merely structural graphs or demographic buckets, but dynamic discursive spaces characterized by "thick" realities \citep{geertz1973interpretation, adilazuarda-etal-2024-towards}. Within these spaces, group identity is not a static property but an active, event-contingent performance, negotiated through linguistically mediated reactions a community adopts to articulate its collective stance.

However, current socio-cultural community alignment evaluations typically rely on coarse abstractions, mapping complex social identities to predefined demographic categories and benchmarking them against aggregated survey data \citep[e.g.][]{santurkar2023opinions, tao2024cultural}. By treating language as a static semantic repository, these approaches capture \textit{what} a community stands for but risk overlooking \textit{how} such stances are pragmatically expressed in real-world discourse. For example, while a community can be semantically categorized as "supportive" of some social value, the manner of that support—whether expressed through weary resignation, sarcastic compliance, or communal solidarity—depicts its social reality. We argue that a community's unique stance is fundamentally encoded in its reaction tones, as these linguistic signals provide a critical dimension for evaluating socio-cultural alignment.

To account for these attitudinal nuances, we shift the evaluation of community-aware socio-cultural alignment from static identity labels to a reaction-centered paradigm and introduce a new benchmark and framework: \logo{C}ommunity-\logo{A}ware \logo{R}eaction \logo{E}valuation (\logo{CARE}). Anchored in the COVID-19 pandemic—a defining period of global fragility encapsulating a dense spectrum of shared uncertainty and localized struggles \citep{bavel2020using, hosseinzadeh2022social}—\logo{CARE} measures whether LLM-simulated discourse aligns with authentic, event-contingent linguistic behaviors observed in online communities. To operationalize this framework, we curated $\sim$10k reactions towards 2,166 real-world news articles and events from this time period, across 207 diverse Reddit communities spanning four continents and ten thematic domains. Consequently, \logo{CARE} provides a rigorous environment for benchmarking how frontier models navigate the "thick" reality of collective human experience.


\begin{figure*}
    \centering
    \includegraphics[width=\textwidth]{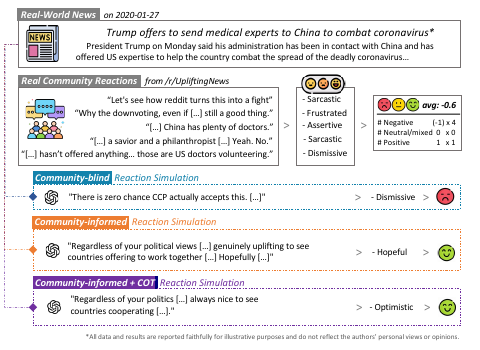}
    \vspace{-0.6cm}
    \caption{Conceptual overview of the \logo{CARE} evaluation framework. While community-informed models attempt to leverage domain context, in this case they over-index on surface-level community hints whereas the authentic community adopts a more realistic and negative stance grounded in the broader context of the event.}
    \label{fig:illustrative}
    \vspace{-0.5cm}
\end{figure*}

Our investigation yields two central insights into community-conditioned social simulation:
\begin{enumerate}
    \item \textbf{Macro-Micro Divergence:} While community conditioning and targeted reasoning significantly enhance macro-level attitudinal and tonal profiling, these gains largely collapse when predicting reactions to specific events. Models anchor to global dataset-level tone distributions and persistent negativity biases rather than identifying instance-level event triggers.
    \item \textbf{Community-Level Volatility:} Prompt conditioning induces highly non-uniform fidelity shifts across sub-communities, where gains in a subset of subreddits are offset by performance drops in others.
\end{enumerate}

We summarize our primary contributions as:
\begin{itemize}
    \item We introduce \logo{CARE}, a reaction-centered diagnostic framework and dataset comprising 9,947 authentic reactions across 207 Reddit communities toward 2,166 real-world news events.
    \item We establish a two-level hierarchical taxonomy that maps 362 fine-grained tone descriptors into 141 L2 synonym groups grounded in Speech Act Theory.
    \item We demonstrate the structural limits of current community-conditioning paradigms, highlighting the critical bottleneck between macro-level profile approximation and micro-level event reaction accuracy.
\end{itemize}
\vspace{-0.3cm}
\section{Related Work}
\vspace{-0.3cm}
While the literature on large language model (LLM) alignment is vast, we constrain our focus to the two domains most critical to our framework: the evaluation of socio-cultural alignment and the application of lexicon-based analysis to social meaning. Broad overviews of general alignment paradigms and safety mechanisms are extensively covered in recent surveys \citep{shen2023large, sucholutsky2023getting, lu2025alignment}.

To evaluate whose values models actually reflect, prompt-based steering has emerged as a highly effective foundational tool for measurement, allowing researchers to probe model biases by conditioning generations on specific demographic or ideological personas \citep{santurkar2023opinions, he2024whose, wu2025axbench, jahanparast2026opinions, occhipinti2026llmasademographic, wang2026siliconsampling}. 
Building on these measurement paradigms, the field has increasingly shifted toward assessing socio-cultural alignment \citep{adilazuarda-etal-2024-towards, chiu-etal-2025-culturalbench, liu-etal-2025-culturally}. Many of these foundational approaches quantify alignment by benchmarking model generations against predefined cultural dimensions and established sociological surveys \citep{santurkar2023opinions, tao2024cultural, masoud-etal-2025-cultural, rao-etal-2025-normad, sukiennik2025evaluation}. However, recognizing that static labels and closed-style evaluations often fail to capture the dynamic reality of these social groups \citep{hovy-yang-2021-importance, rottger-etal-2024-political, kabir2025break, shen-etal-2025-revisiting}, subsequent work such as \cite{liu2025cultural} explores dynamic cultural adaptation through role-play-inspired simulation and cultural learning pipelines. 

Closest to our objective is the emerging subfield of community-level alignment \citep{shi-etal-2024-culturebank, chu2025improving, lin2026communitybench}. While these approaches capture broad community archetypes \citep{prinster2024community}, modeling the authentic, event-contingent reactions that characterize how real-world online groups navigate complex social shifts remains an open frontier. To bridge this realism gap, we draw on the long-standing tradition of lexicon-based analysis in NLP, which offers a robust, interpretable mechanism for evaluating text beyond surface-level generation. Extensive prior work has utilized linguistic schemata to tackle a wide array of foundational NLP and computational linguistic tasks \citep[e.g.][]{stolcke-etal-2000-dialogue, pennebaker2007linguistic, ding2008holistic}, while more recently, lexicon-driven approaches have evolved to tackle complex tasks that probe deeper into social meaning \citep{havaldar-etal-2024-building, huang2024characterizing} and inspired this work.
\section{Corpus and Dataset Construction}
The \logo{CARE} raw corpus comprises all posts on reddit.com of year 2020 containing either of the query words "coronavirus" or "covid". We restrict content to English to isolate the dataset from multilingual effects during evaluation.  

Table \ref{tab:dataset_stats} reports dataset statistics for each processing stage detailed below.
\begin{table*}[htbp]
\centering
\resizebox{\textwidth}{!}{%
\begin{tabular}{@{}rrrrrrr@{}}
\toprule
\textbf{Processing Stage} & \textbf{\# Subreddits} & \textbf{\# Articles} & \textbf{\# Keyphrases} & \textbf{\# Posts} & \textbf{\# Reactions} & \textbf{Avg. React. Len.} \\ \midrule
(Initial Corpus) & 4,206 & 117,747 & 199,399 & 141,487 & 651,433 & 40.35 \\
Subreddit Selection & 207 & 99,531 & 170,055 & 117,478 & 546,357 & 40.68 \\
Keyphrase Samp. & 207 & 57,019 & 94,431 & 67,654 & 315,977 & 40.03 \\
Temporal Samp. & 207 & 2,166 & 8,431 & 2,194 & 9,947 & 42.43 \\ \bottomrule
\end{tabular}%
}
\caption{Unique counts and average word lengths of the \logo{CARE} corpus across all processing stages. \looseness=-1}
\label{tab:dataset_stats}
\end{table*}

\subsection{From Raw Data to Reaction Corpus}

To ensure that the collected data reflects community reactions to external events, only posts hyperlinked to external news articles are considered. They also must have exhibited sufficient community engagement (received at least $10$ comments). Since nested comments are more likely to deviate from the original news content, we treat top-level comments directly replying to the posts as news reactions and rank them by their up-vote scores. Up to 5 highest-ranked top-level comments within each post are selected to represent the community reactions towards the posted news event. Together with other community and news information, they form the initial \logo{CARE} corpus.

\subsection{Data Sampling for Evaluation}

Due to the scale and heterogeneity of the initial corpus, the final evaluation set is curated via a structured down-sampling process designed to best maintain cross-community coverage and data representativeness. This process includes 3 consecutive stages:

\paragraph{Community Selection} We select subreddits that (1) sustained active participation in COVID-19 discourse, and (2) exhibited overlapping engagement over news articles. This overlap ensures that selected communities operate within a broadly shared informational landscape, grounding the resulting corpus in a cohesive domain of discourse.

In terms of implementation, 256 subreddits with at least two COVID-19–related posts per month for nine or more months are retained. In parallel, all subreddits are ranked by the number of shared articles they reacted to, and the top 300 are selected. Intersecting the two sets yields 207 subreddits, which constitute our final community pool. Additionally, each selected community is manually assigned a topical category and a full geographic path description by the authors.\footnote{Figure \ref{fig:dataset_coverage} in Appendix \ref{sec:appendix_corpus} visualizes the topical x geographical distribution over all 207 subreddits.}

\paragraph{Keyphrase-based News Sampling} To guide news article sampling, we apply an unsupervised keyphrase extraction and ranking algorithm \citep{boudin2018unsupervised} to each news article. Keyphrases are ranked by a saliency score designed to favor phrases that appear both frequently and broadly across the selected subreddits. Formally, let $K \in \mathbb{R}^{n \times m}$ denote the keyphrase--subreddit count matrix, where $n$ is the number of keyphrases and $m$ the number of selected subreddits. The resulting salience vector $\mathbf{s} \in \mathbb{R}^n$, defined over keyphrases, is given by
\[
\mathbf{s}
=
\log(1 + K\mathbf{1}_m)
\odot
\tfrac{1}{m}\,\mathds{1}[K > 0]\mathbf{1}_m,
\]
where $\mathbf{1}_m$ is an $m$-dimensional all-ones vector, $\mathds{1}[\cdot]$ denotes the indicator function, and $\odot$ indicates element-wise multiplication. The first term captures aggregated keyphrase frequency across subreddits, while the second measures the fraction of subreddits in which a keyphrase appears at least once. Figure \ref{fig:keyphrase_sampling} presents a distributional analysis of salience scores, raw frequencies, and sparsity levels, illustrating that the heavy-tailed patterns motivating our formulation are preserved. 

After a manual check on the top 300 keyphrases, 28 semantically ambiguous terms are removed, rendering a sample set of 272 keyphrases. Posts linked to news articles with at least one of the sampled keyphrases are retained for the final round of data sampling.

\paragraph{Temporal Sampling} To ensure temporal and community balance under budget constraints, we randomly select up to one post per community per month of the year 2020. This yields the final evaluation set consisting of 2,194 posts and 9,947 reactions across 207 Reddit communities. 
\section{Evaluation Framework}

\subsection{Simulation Setup}
We evaluate models under three controlled prompting conditions that progressively incorporate community context and deliberate pragmatic reasoning. For each community-post pair, exactly one simulated reaction comment is elicited per prompt setting and evaluated against the corresponding set of authentic community reactions:\footnote{Complete prompt templates are  available in Appendix \ref{sec:appendix_prompt}.}
\begin{itemize}
    \item \textbf{Community-Blind} Models receive only the news article headline and content alongside standard instruction to simulate a Reddit-style news comment, with no information regarding the originating subreddit.
    \item \textbf{Community-Informed} The prompt template is augmented with community-specific metadata, explicitly specifying the target subreddit identifier (e.g., \textit{/r/UpliftingNews}).
    \item \textbf{Community-Informed-CoT} In addition to the subreddit identifier and article text, the prompt incorporates the subreddit's official self-description and mandates a step-by-step chain-of-thought (CoT) reasoning process—analyzing community values, news content nuances, and contextual factors—before producing the final comment.
\end{itemize}

\subsection{Linguistic Schema}
Central to our evaluation framework is a reaction-level linguistic schema that characterizes responses in terms of reaction tones and attitudes, providing an interpretable abstraction from raw text to comparable evaluative signals.

\paragraph{Defining Reaction Tone}
Building on recent work that uses descriptive adjectives to characterize conversational tone divergences between humans and LLMs \citep{huang2024characterizing}, our definition of tone is inspired by Speech Act Theory \citep{austin1975things, searle1979expression}: rather than analyzing the literal utterance (\textbf{locutionary act}) or the resulting effect (\textbf{perlocutionary effect}), we operationalize tone via the speaker's orientation on the \textbf{illocutionary} level. By focusing on the underlying intent or social action of the speaker, we move beyond surface-level semantics to capture the specific pragmatic strategies through which a community's collective stance is actively voiced.

\paragraph{Mapping Tones to Attitudes}
Attitude serves as a higher-level abstraction over tone, capturing both stance (support, opposition, or neutrality toward a target) and valence (positive, negative, or neutral/mixed evaluation). While any evaluative act conceptually contains both a positive and a negative pole, tone provides the observable signal that determines which orientation is actively expressed. Consequently, our coarse attitude labels (\textit{positive}, \textit{negative}, \textit{neutral/mixed}) reflect the explicit directional orientation conveyed toward the referenced event.

\subsection{Annotation Pipeline and Taxonomy Derivation}
We establish ground-truth and simulated labels via a human-AI collaborative pipeline, moving from single-utterance tone annotation to hierarchical taxonomy construction.

\paragraph{Reaction Tone Annotation}
Tone annotation was executed via batched LLM inference, guided by a pilot study to establish annotation quality. To ensure cross-community robustness, we sampled representative reactions across $207$ subreddits and benchmarked six annotation setups. The optimal configuration using GPT-5 achieved an acceptance rate of $96.6\%$ on manual validation (details in Appendix \ref{sec:appendix_iaa}) and was deployed to label both authentic community reactions and model simulations.

\paragraph{Hierarchical Tone Taxonomy}
To operationalize this linguistic schema without data sparsity or lexical redundancy, we organize tone words into a two-level hierarchy. Level 1 (L1) partitions descriptors into broad pragmatic categories aligned with attitude directions. Level 2 (L2) creates mutually exclusive sub-clusters within each L1 category, grouping tight synonym sets while isolating standalone pragmatic acts. This structure creates a clear synergy by mapping specific speech acts (L2) through pragmatic strategies (L1) up to overarching stance and valence (attitude). Across our dataset, the taxonomy consolidates 362 distinct ground-truth and simulated tone descriptors into 141 L2 synonym groups. The taxonomy was co-developed through an iterative collaboration between the human authors and multiple state-of-the-art LLMs. A full listing is available in Appendix \ref{sec:appendix_taxonomy}.

\subsection{Alignment Metrics}
We quantify alignment between simulated comments and authentic community reactions at both the attitude and tone levels, balancing instance-level accuracy with corpus-level distributional consistency.

\paragraph{Attitude Alignment}
Attitude alignment measures how accurately simulated reactions replicate community valence and stance. We evaluate instance-level accuracy via Attitude Hit Rate ($\text{AttHit}$), defined as the proportion of simulations whose derived attitude matches the ground-truth community reaction. To capture fine-grained deviations, we compute Root Mean Squared Error ($\text{RMSE}$) over continuous community attitude scores, alongside Mean Error ($\text{ME}$) partitioned into negative and non-negative subsets to diagnose directional overestimation or underestimation bias. Rank consistency across posts is measured using Spearman's rank correlation ($\rho$).

\paragraph{Tone Alignment}
Tone alignment measures pragmatic fidelity across both raw lexical descriptors and L2 synonym groups. At both levels, we report Exact Match to measure whether a generated tone falls within the set of ground-truth community reactions for a given post. To assess macro-level pragmatic distribution, we calculate the Jensen-Shannon Divergence between the generated and authentic tone distributions across the dataset. Lexical and cluster diversity are tracked via total unique word and group counts.

\section{Evaluation Results And Analysis}
\label{sec:experiments}

Our empirical evaluation follows a structured progression: we first present the aggregate, macro-level community profiling results, which initially suggest substantial alignment gains from community conditioning. We then contrast these macro results with instance-level, event-contingent metrics, revealing a fundamental collapse in micro-level reaction accuracy. Finally, we examine sub-community transition patterns to uncover the underlying representational trade-offs and disparities induced by prompt conditioning. To track simulation behavior across consecutive model generations, all simulations are conducted using two flagship releases from the same model family: Gemini 2.5 Pro and Gemini 3.1 Pro Preview.

To ground these model evaluations against non-parametric references, we benchmark model predictions against reference baselines across the full space of $362$ taxonomy tones. The Random reference samples uniformly from all $362$ classes, establishing a strict chance lower bound. The Top baseline assigns the tone of the single most upvoted authentic reaction, capturing the most popular community response. We evaluate four mixture configurations—100:0, 50:50, 25:75, and 0:100—establishing a calibrated performance gradient from uniform chance to dominant community consensus.

\subsection{Macro-Level Profile Alignment}
\label{sec:macro_alignment}

We begin our empirical evaluation by examining whether language models can reproduce broad community profiles. By aggregating individual reaction tones and attitudes into community-level summary distributions across all $N = 207$ subreddits, we evaluate macro fidelity across two core dimensions: community attitude dynamics and tonal structure.

\begin{figure*}
    \centering
    \includegraphics[width=\textwidth]{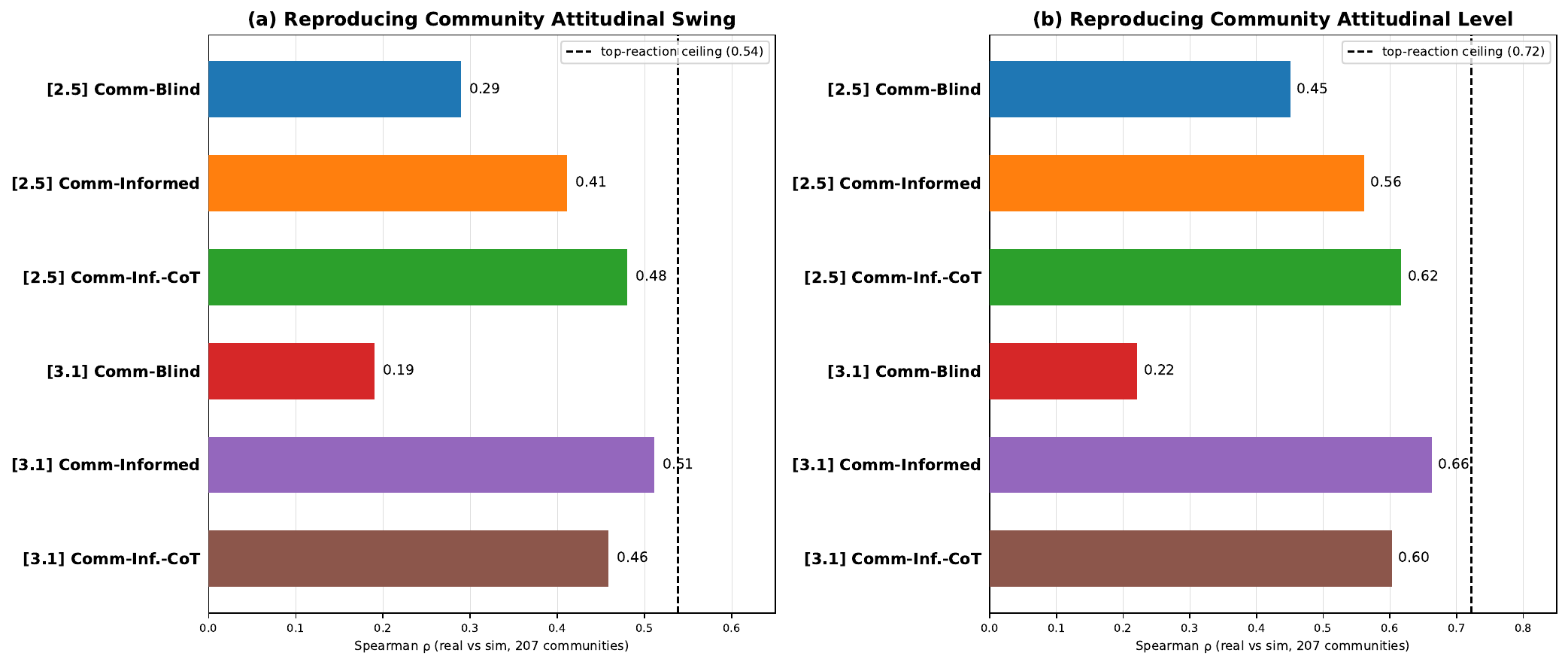}
    \caption{Macro-level attitudinal profile alignment.}
    \label{fig:attitude_corr}
\end{figure*}
\paragraph{Attitudinal Dynamics: Level and Swing}
To capture community attitudinal landscapes, we compute two aggregate metrics per subreddit $c$: Attitudinal Level (mean attitude score, $\mu_c$) and Attitudinal Swing (attitude dispersion, measured via standard deviation $\sigma_c$). We then report the Spearman rank correlation ($\rho$) between real and simulated community vectors ($\mu$ or $\sigma$ for all $c$'s) (Figure~\ref{fig:attitude_corr}).

While unconditioned baselines (Comm-Blind) struggle most to simulate community attitude, providing explicit community context (Comm-Informed) yields dramatic fidelity leaps, particularly for Gemini 3.1 Pro. Gemini 2.5 Pro displays a steady, monotonic progression across prompt conditions, reaching its peak under Chain-of-Thought prompting (Comm-Inf.-CoT). Notably, while CoT is necessary for Gemini 2.5 Pro to fully synthesize community context, adding CoT to Gemini 3.1 Pro causes a slight performance regression, suggesting that extensive step-by-step reasoning can introduce slight normative drift away from direct community priors.

\begin{figure*}
    \centering
    \includegraphics[width=\textwidth]{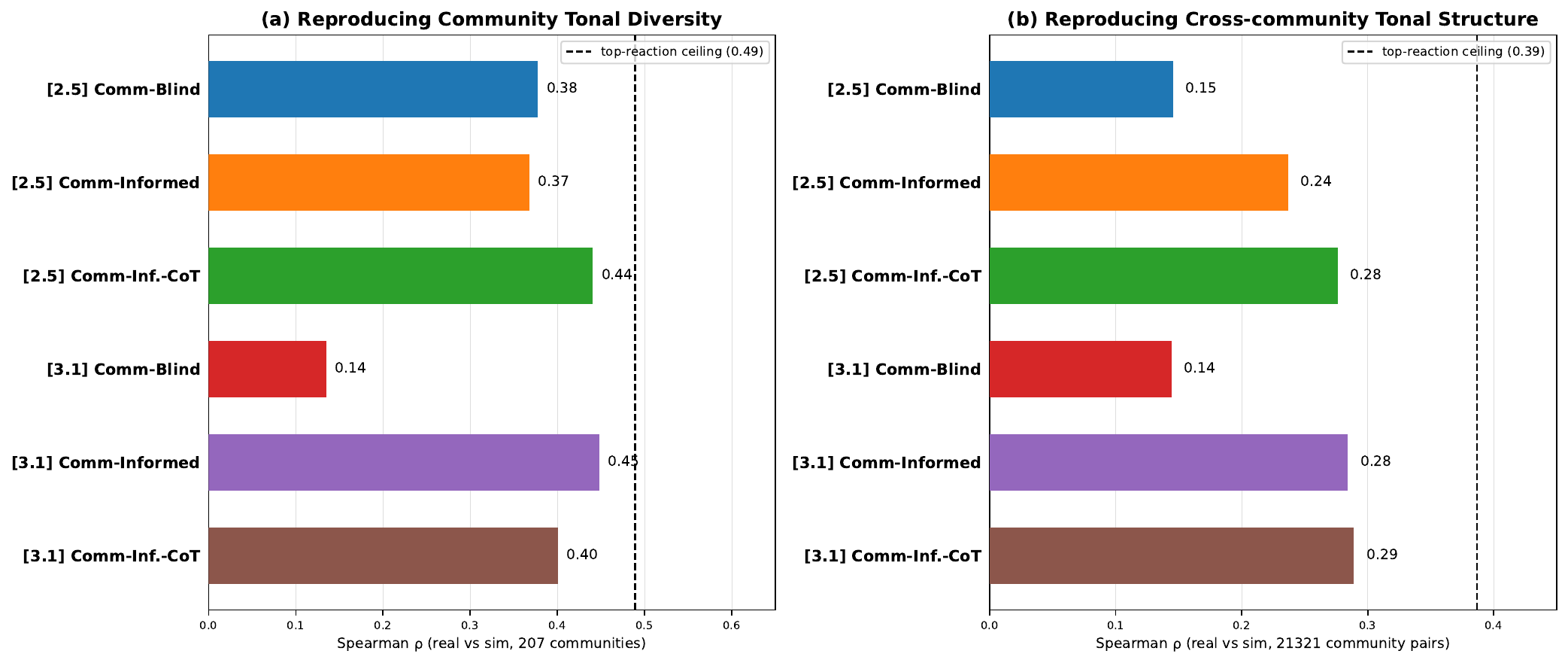}
    \caption{Macro-Level Tonal Profile Alignment.}
    \label{fig:tone_corr}
\end{figure*}
\paragraph{Tonal Diversity and Cross-Community Structure}
A parallel trend emerges when evaluating communicative tone across communities (Figure~\ref{fig:tone_corr}). For intra-community \textit{Tonal Diversity} (measured via L2-grouped entropy), community conditioning elevates Gemini 3.1 Pro drastically compared with the blind setting, while Gemini 2.5 Pro reaches its peak via CoT. Furthermore, when assessing \textit{Cross-Community Tonal Structure}---evaluated as the Spearman correlation across all $\binom{207}{2} = 21,321$ pairwise community Jensen-Shannon distances--we observe a clear upward trend for both models.

Taken together, these macro-level correlations seem to suggest that community conditioning and explicit reasoning successfully steer LLMs toward more community-aware profiles. However, as we demonstrate in the following section, evaluating fidelity solely through aggregate distributions creates a deceptive veneer of success: when we evaluate LLM simulations at a more fine-grained, event-contingent post level, these macro gains largely collapse.

\subsection{Micro-Level Event-Contingent Reaction Collapse}
\label{subsec:micro_collapse}

\begin{table}[htbp]
  \centering
  \resizebox{0.8\textwidth}{!}{%
  \begin{tabular}{ll ccccc}
    \toprule
    \textbf{Model} & \textbf{Style / Ratio} & \textbf{AttHit} $\uparrow$ & \textbf{RMSE} $\downarrow$ & \textbf{$\text{ME}_{\text{neg}}$} & \textbf{$\text{ME}_{\text{non-neg}}$} & \textbf{Spearman $\rho$} $\uparrow$ \\
    \midrule
    \multirow{3}{*}{Gemini 2.5 Pro} 
      & Comm-Blind        & 0.63 & 0.70 & -0.06 & -0.51 & 0.27 \\
      & Comm-Informed     & 0.62 & 0.71 & -0.01 & -0.35 & 0.31 \\
      & Comm-Inf.-CoT     & 0.60 & 0.72 & \phantom{-}0.03 & -0.34 & 0.30 \\
    \midrule
    \multirow{3}{*}{Gemini 3.1 Pro Prev.} 
      & Comm-Blind        & 0.59 & 0.76 & \phantom{-}0.00 & -0.65 & 0.13 \\
      & Comm-Informed     & 0.63 & 0.66 & -0.06 & -0.37 & 0.34 \\
      & Comm-Inf.-CoT     & 0.62 & 0.67 & -0.02 & -0.42 & 0.31 \\
    \midrule
    \multirow{4}{*}{\shortstack[l]{References\\(Random : Top)}} 
      & 100 : 0           & 0.40 & 0.95 & \phantom{-}0.46 & -0.41 & -0.01 \\
      & 50 : 50           & 0.54 & 0.77 & \phantom{-}0.21 & -0.29 & 0.28 \\
      & 25 : 75           & 0.62 & 0.66 & \phantom{-}0.08 & -0.11 & 0.43 \\
      & 0 : 100           & 0.68 & 0.56 & -0.04 & -0.03 & 0.57 \\
    \bottomrule
  \end{tabular}%
  }
    \caption{Attitude-level alignment metrics across models and reference baselines.}
    \label{tab:attitude_metrics_summary}
\end{table}
\paragraph{Attitudinal Stagnation and Directional Bias}
In sharp contrast to macro-level correlation gains, instance-level attitude metrics (Table~\ref{tab:attitude_metrics_summary}) reveal widespread stagnation across prompting strategies. For Gemini 2.5 Pro, $\text{AttHit}$ and $\text{RMSE}$ remain practically unchanged across conditions, hovering near the 25:75 reference benchmark. While Gemini 3.1 Pro exhibits a noticeable jump from Comm-Blind to Comm-Informed, the blind setting suffers from an initial performance slump, and the conditioned prompting seems to recover its alignment back to standard baseline levels before immediately plateauing under CoT. Furthermore, directional error analysis uncovers a persistent negativity bias. While mean error in negatively reacted events remains calibrated near zero across all conditions,\footnote{The signed Mean Error ($\text{ME}$) is inherently susceptible to error neutralization if positive and negative deviations cancel out within a dataset.} models consistently simulate unwarranted negative attitudes for non-negative instances.

\begin{table}[htbp]
  \centering
  \resizebox{\textwidth}{!}{%
  \begin{tabular}{ll cccc cccc}
    \toprule
    & & \multicolumn{4}{c}{\textbf{Exact Tone Metrics}} & \multicolumn{4}{c}{\textbf{Synonym Group Metrics (\textit{syn})}} \\
    \cmidrule(lr){3-6} \cmidrule(lr){7-10}
    \textbf{Model} & \textbf{Style / Ratio} & \textbf{EM} $\uparrow$ & \textbf{JSD} $\downarrow$ & \textbf{$N_{\text{unique}}$} & \textbf{\% Sarcastic} & \textbf{$\text{EM}_{\text{syn}}$} $\uparrow$ & \textbf{$\text{JSD}_{\text{syn}}$} $\downarrow$ & \textbf{$N_{\text{groups}}$} & \textbf{\% $\text{Sarcastic}_{\text{syn}}$} \\
    \midrule
    (Real Reactions) & -- & 1.00 & 0.00 & 328 & 22.22 & 1.00 & 0.00 & 136 & 23.38 \\
    \midrule
    \multirow{3}{*}{Gemini 2.5 Pro} 
      & Comm-Blind        & 0.36 & 0.42 & 128 & 37.33 & 0.41 & 0.36 & 74 & 37.92 \\
      & Comm-Informed     & 0.35 & 0.39 & 139 & 30.22 & 0.40 & 0.32 & 81 & 30.67 \\
      & Comm-Inf.-CoT & 0.31 & 0.42 & 139 & 22.15 & 0.37 & 0.36 & 73 & 22.52 \\
    \midrule
    \multirow{3}{*}{Gemini 3.1 Pro Prev.} 
      & Comm-Blind        & 0.31 & 0.41 & 112 & 35.14 & 0.36 & 0.37 & 64 & 36.46 \\
      & Comm-Informed     & 0.34 & 0.41 & 119 & 28.40 & 0.42 & 0.36 & 68 & 29.40 \\
      & Comm-Inf.-CoT & 0.31 & 0.42 & 130 & 22.52 & 0.38 & 0.38 & 70 & 23.38 \\
    \midrule
    \multirow{4}{*}{\shortstack[l]{References\\(Random : Top)}} 
      & 100 : 0 & 0.01 & 0.70 & 362 &  0.27 & 0.04 & 0.59 & 141 &  0.50 \\
      & 50 : 50 & 0.51 & 0.40 & 351 & 12.35 & 0.53 & 0.31 & 141 & 12.85 \\
      & 25 : 75 & 0.76 & 0.27 & 321 & 17.50 & 0.76 & 0.20 & 131 & 18.14 \\
      & 0 : 100 & 1.00 & 0.17 & 206 & 23.61 & 1.00 & 0.12 & 106 & 24.43 \\
    \bottomrule
  \end{tabular}%
  }
    \caption{Tone-level alignment metrics across models and reference baselines.}
    \label{tab:tone_metrics_summary}
\end{table}

\paragraph{Tonal Stagnation and Non-Micro Calibration}
A similar stagnation pattern appears in tone evaluation (Table~\ref{tab:tone_metrics_summary}) except that we do not observe an obvious performance slump in Gemini 3.1 Pro over Comm-Blind setting, leading to its slight improvement after community information injection (Comm-Informed). Interestingly, we find CoT prompt successfully suppresses the unconditioned sarcasm inflation distributionally, pulling global sarcasm output into near-perfect alignment with authentic human baselines. However, with multiple drops in instance-level metrics observed, we suspect that this calibration operates more at the macro level---knowing how much sarcasm to generate overall across a dataset without identifying which specific event could trigger a sarcastic community reaction.

\subsection{Dynamics of Community-Conditioned Tone Transitions}
\label{subsec:disparities}

\begin{figure*}
    \centering
    \includegraphics[width=\textwidth]{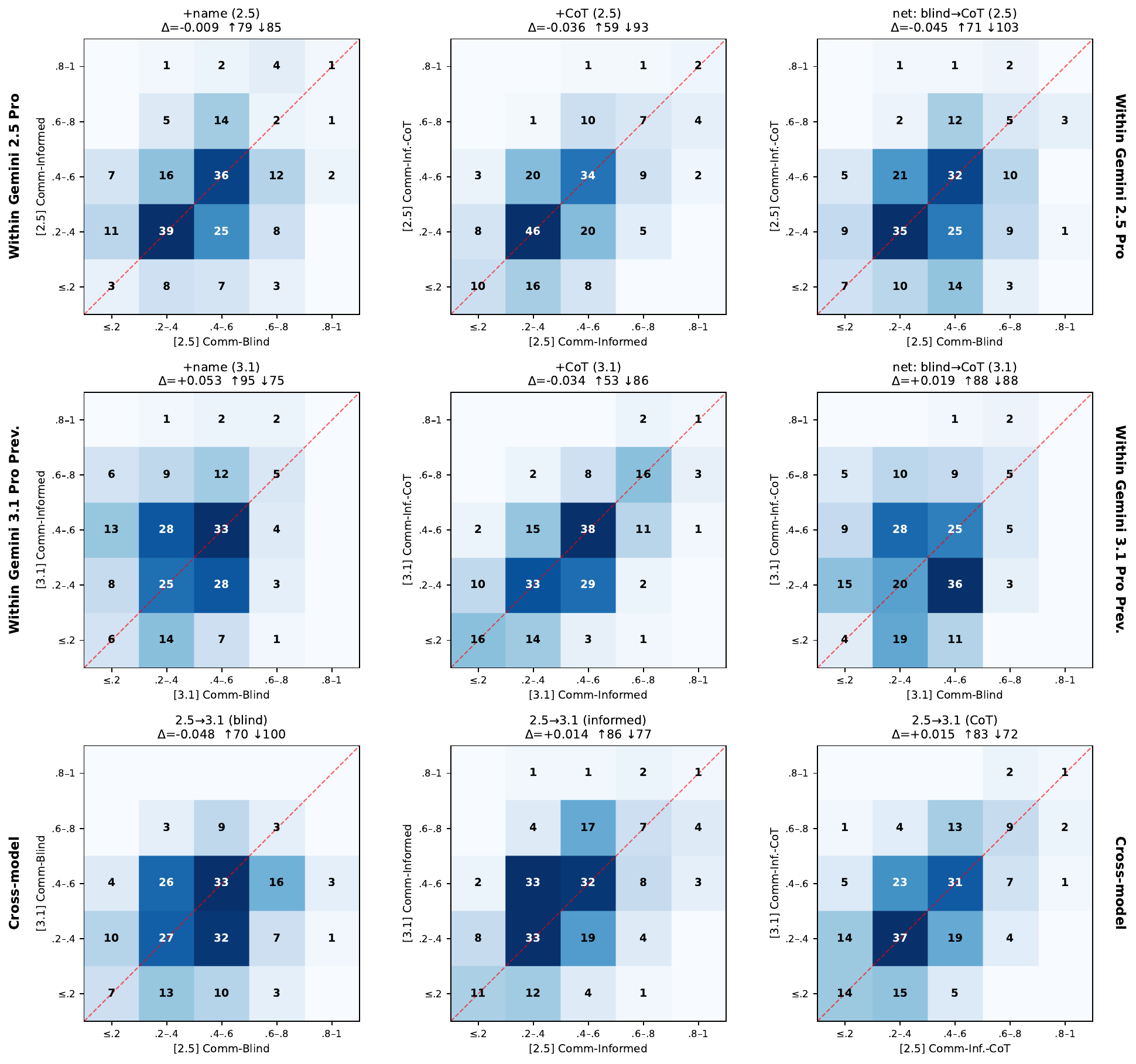}
    \caption{Community-level tonal shifts in simulation fidelity under different settings.}
    \label{fig:tone_shifts}
\end{figure*}

To further examine how tone predictions change under community conditioning, we analyze community-level transitions in L2 tone exact match scores using aggregated heatmaps (Figure \ref{fig:tone_shifts}). Each heatmap visualizes how communities move between $\text{EM}_{syn}$ score intervals when transitioning between different model and prompt style settings, with diagonal mass indicating stability and off-diagonal mass indicating shifts in tone-level fidelity. We find:

\begin{itemize}
    \item \textbf{Identifier Transition (\textit{+name}):} Adding community identifiers improves exact tone match for 79 subreddits in Gemini 2.5 Pro, while degrading 85 subreddits. For Gemini 3.1 Pro, 95 subreddits improve while 75 degrade.
    \item \textbf{Reasoning Transition (\textit{+CoT}):} Incorporating explicit community context and CoT reasoning yields 59 gains vs. 93 losses in Gemini 2.5 Pro; 53 gains vs. 86 losses in Gemini 3.1 Pro.
    \item \textbf{Net Trajectory:} Over the complete trajectory from Comm-Blind to Comm-Inf.-CoT, 71 subreddits gain tone fidelity while 103 suffer net degradation in Gemini 2.5 Pro; 88 gains vs. 88 losses in Gemini 3.1 Pro.
\end{itemize}

These heterogeneous transition patterns naturally raise a follow-up question: which communities are affected by community conditioning, and in what ways? In particular, one may ask whether the observed gains and losses concentrate around specific types of communities, or whether certain groups are systematically favored or disadvantaged. Given the authors’ limited domain expertise across the full range of 207 communities studied, we refrain from attributing the observed inequalities to specific cultural, topical, or geographic factors. We view this as an important open question and welcome further investigation by domain experts to better understand the fairness implications of community-conditioned modeling.

\subsection{Interpreting Community Alignment Via and Beyond Metrics}
\label{subsec:beyond_metrics}

Taken together, our analysis highlights the inherently multi-faceted nature of community-aware behavior and the limits of interpreting it through any single metric. Attitude- and tone-level evaluations consistently show that different metrics capture different aspects of model response, including instance-level fidelity, distributional resemblance, and community-specific variability. These divergences are not incidental but structurally informative. Holding these perspectives together allows changes to be interpreted in terms of how behavior shifts, redistributes, or concentrates across instances and communities, rather than whether a model simply improves or degrades by a particular score. In this sense, quantitative metrics function less as final arbiters of community awareness and more as instruments for probing a complex, socially grounded form of model behavior.
\section{Conclusion}

In this work, we introduce \logo{CARE}, a reaction-centered framework designed to evaluate how faithfully large language models simulate the "thick" realities of online communities. Our analysis reveals a persistent "realism gap" in current alignment paradigms. Explicitly conditioning models on community identity does not inherently yield uniform improvements in simulation fidelity. Ultimately, \logo{CARE} functions as a diagnostic tool showing that static community prompting fails to capture dynamic event reactions, highlighting the need for more dynamic, event-aware alignment strategies in social and community simulation.

\clearpage
\section*{AI Use Statement}
We disclose the use of Large Language Models (LLMs) across the following phases of this work:

\begin{itemize}
    \item \textbf{Writing Assistance, Drafting, and Polishing:} LLMs were used to assist in drafting, rephrasing, and polishing prose, improving grammatical clarity, and structuring select sections of the manuscript. All text was strictly reviewed, edited, and validated by the human authors to ensure scientific accuracy and original framing.
    
    \item \textbf{Research Ideation and Execution:} LLMs were utilized during experimental execution to co-develop the two-level illocutionary tone taxonomy and infer preliminary tone-to-attitude mappings, which were then qualitatively audited and finalized by the authors. Our codebase is developed with the assistance of Claude Code.
    
    \item \textbf{Synthetic Dataset Generation:} Generative language models (Gemini 2.5 Pro and Gemini 3.1 Pro Preview) were used to generate simulated community reactions across the 207 subreddits and 2,194 posts evaluated under our benchmark prompting paradigms. GPT-5 was also employed as an automated annotator for categorizing fine-grained communicative tones across human and simulated reactions. 
    
\end{itemize}

\bibliography{reference}
\bibliographystyle{iclr2027_conference}
\appendix
\setcounter{figure}{0} 
\renewcommand{\thefigure}{\thesection\arabic{figure}} 

\section{Additional Corpus Details}
\label{sec:appendix_corpus}

Here we provide additional figures regarding the composition and sampling of the CARE corpus. 

\begin{figure*}
\centering
\includegraphics[width=\textwidth]{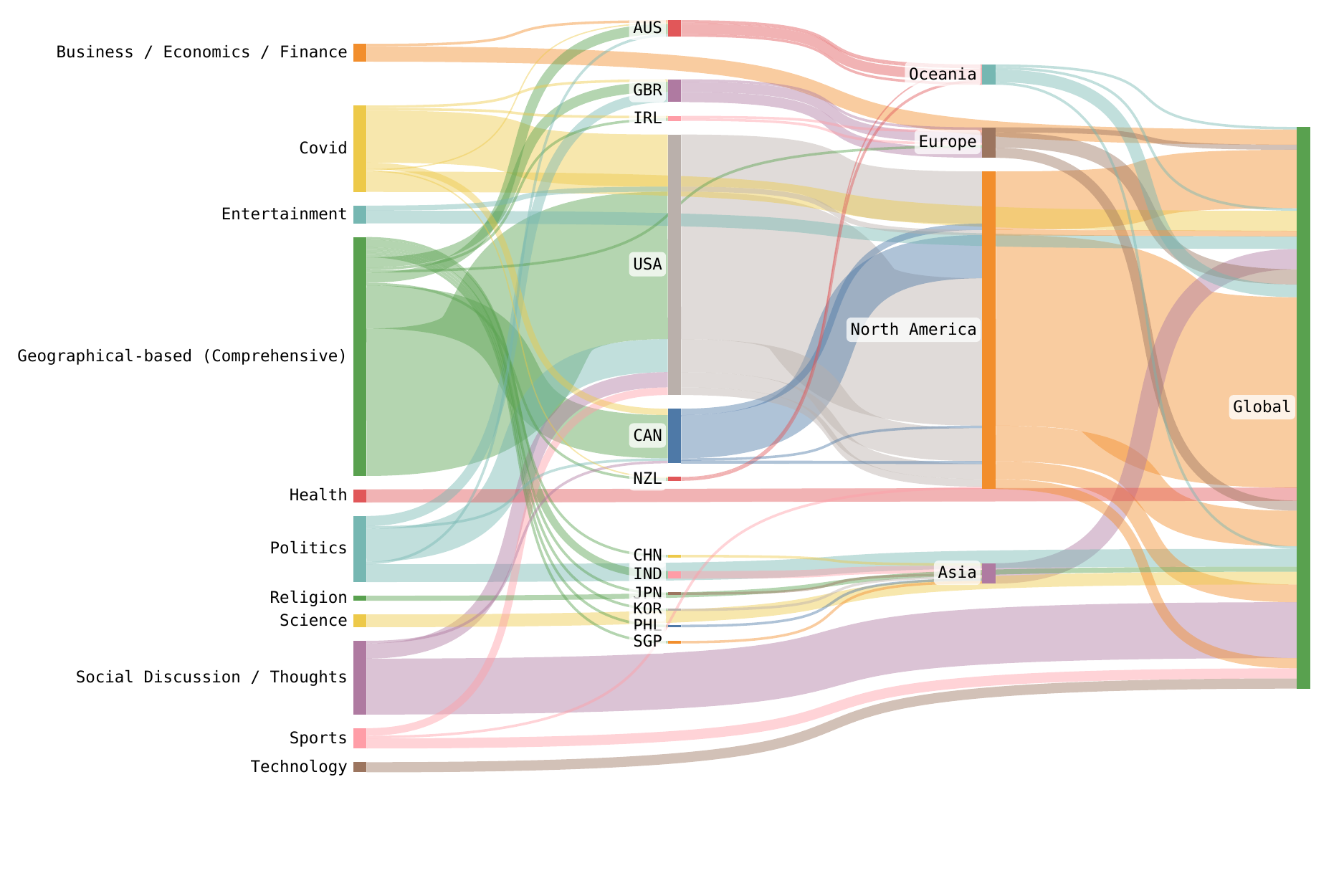}
\vspace{-1.5cm}
  \caption{Sankey diagram illustrating the joint distribution of primary topics and down-to-country-level geographic scopes across the 207 selected communities. Although this visualization aggregates flows at or above the country level, our dataset's geographic annotations are more granular and diverse, scaling from unconstrained global discourse (e.g., \texttt{GLOBAL}) down to specific administrative and city levels (e.g., \texttt{GLOBAL-NORTH AMERICA-USA-CA-LosAngeles}), as in-depth as necessary.}
  \label{fig:dataset_coverage}
\end{figure*}

Figure \ref{fig:dataset_coverage} visualizes the joint distribution of primary discussion topics and macro-level geographic scopes across the 207 selected Reddit communities, highlighting the thematic and regional diversity of our corpus. 

\begin{figure}
\centering
\includegraphics[width=\textwidth, height=0.9\textheight, keepaspectratio]{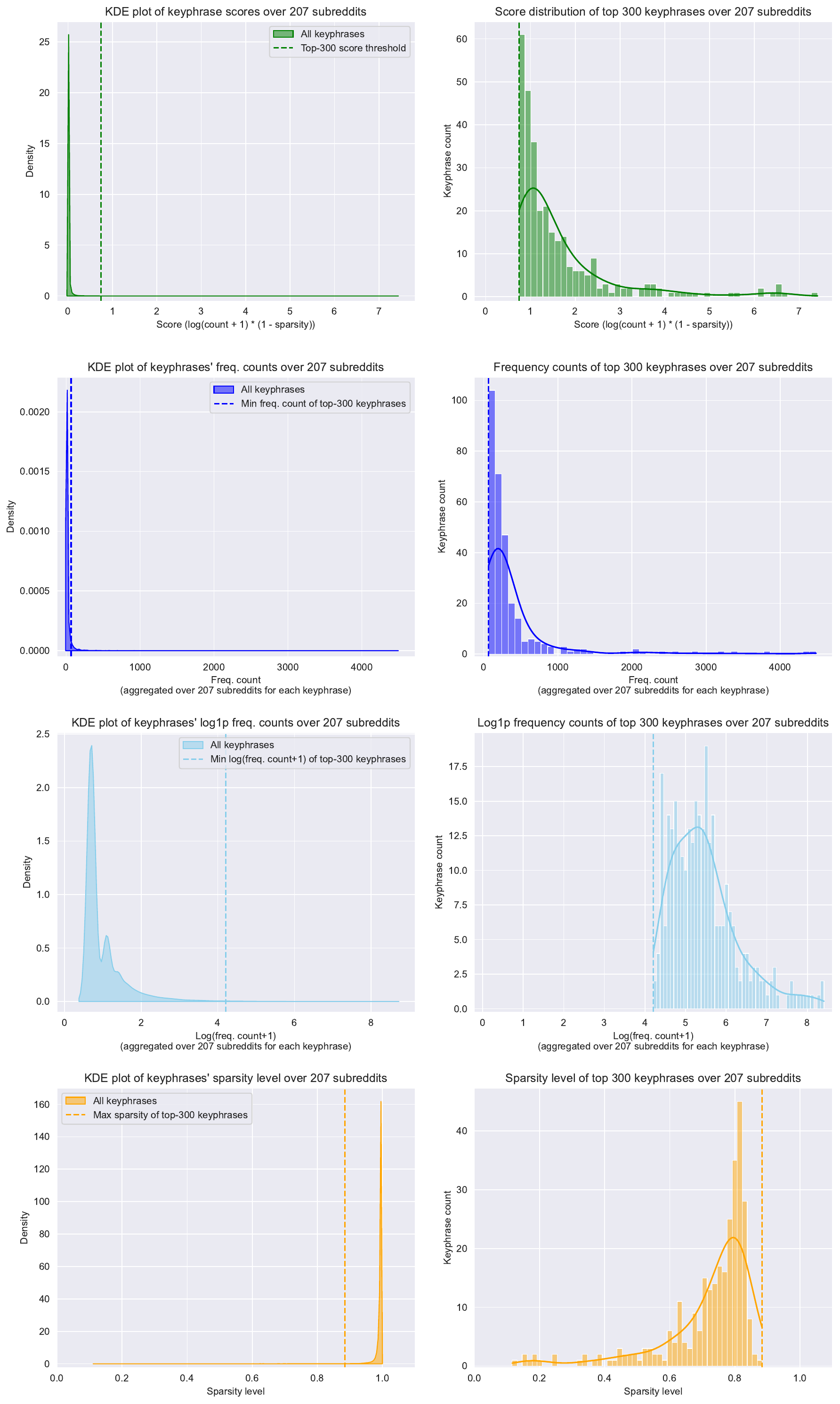}
\caption{Distributional analysis of candidate keyphrases across the 207 selected subreddits.}
\label{fig:keyphrase_sampling}
\end{figure}

Figure \ref{fig:keyphrase_sampling} provides a detailed distributional analysis of the keyphrase sampling process, illustrating how our salience scoring formulation preserves heavy-tailed frequency patterns while effectively filtering for phrases that are broadly discussed across the community pool.

\section{Hierarchical Tone Taxonomy Details}
\label{sec:appendix_taxonomy}

The full tone taxonomy is listed out in Table \ref{tab:tone_taxonomy_negative}-\ref{tab:tone_taxonomy_mixed}.

\begin{xltabular}{\textwidth}{@{} >{\raggedright\arraybackslash}p{0.32\textwidth} X @{}}
\caption{Hierarchical Tone Taxonomy: Negative Attitude Domain} \label{tab:tone_taxonomy_negative} \\
\toprule
\textbf{L2 Subgroup} & \textbf{Tone Synonyms} \\
\midrule
\endfirsthead
\caption[]{Hierarchical Tone Taxonomy: Negative Attitude Domain (Continued)} \\
\toprule
\textbf{L2 Subgroup} & \textbf{Tone Synonyms} \\
\midrule
\endhead
\bottomrule
\endlastfoot
\multicolumn{2}{@{}l@{}}{\textbf{Anger \& Hostility}} \\
\cmidrule(r){1-2}
  Overt Rage                       & angry, outraged, infuriated, furious \\
  Irritation \& Impatience         & frustrated, irritated, exasperated, impatient, annoyed \\
  Resentful Indignation            & resentful, indignant, bitter \\
  Militant Aggression              & combative, confrontational, violent, hostile, militant, aggressive \\
  Acrimonious Attack               & acerbic, scathing, vitriolic, caustic \\
  Harsh Bluntness                  & curt, abrasive, strident \\
  Spiteful Retaliation             & vindictive, vengeful, spiteful \\
  Provocative Incitement           & inflammatory, incendiary, provocative \\
  Bias \& Hatred                   & hateful, racist, xenophobic \\
  Vulgarity                        & profane \\
\addlinespace[1em]
\multicolumn{2}{@{}l@{}}{\textbf{Contempt Mockery \& Wit}} \\
\cmidrule(r){1-2}
  Arrogant Patronage               & patronizing, condescending \\
  Disgust \& Scorn                 & contemptuous, disdainful, disgusted, scornful \\
  Derisive Ridicule                & mocking, derisive, derogatory \\
  Sharp Sarcasm                    & snarky, sarcastic \\
  Sneaky Insults                   & insulting, snide, insinuating \\
  Taunting \& Triumph              & gloating, taunting \\
  Playful Ribbing                  & cheeky, teasing \\
  Irreverent Levity                & flippant, facetious \\
  Dark \& Critical Humor           & satirical, sardonic \\
  Dry \& Paradoxical Wit           & ironic, wry \\
  Self Targeted Wit                & self-deprecating \\
  Active Dismissal                 & dismissive \\
\addlinespace[1em]
\multicolumn{2}{@{}l@{}}{\textbf{Judgment \& Disapproval}} \\
\cmidrule(r){1-2}
  Accusation \& Condemnation       & accusatory, condemnatory, punitive \\
  Reproof \& Correction            & corrective, admonitory, admonishing, scolding \\
  Critical Evaluation              & judgmental, disapproving, critical \\
  Moralizing Instruction           & didactic \\
  Argumentative Militancy          & argumentative, contrarian, litigious, polemical \\
  Imperative Pressure \& Ranting   & demanding, bombastic \\
\addlinespace[1em]
\multicolumn{2}{@{}l@{}}{\textbf{Doubt Confusion \& Shock}} \\
\cmidrule(r){1-2}
  Skepticism \& Suspicion          & incredulous, suspicious, questioning, skeptical \\
  Disorientation \& Bafflement     & bewildered, confused, bemused, baffled \\
  Shock \& Stunned Silence         & stunned, appalled, shocked \\
  Paranoia \& Secrecy              & paranoid, conspiratorial \\
\addlinespace[1em]
\multicolumn{2}{@{}l@{}}{\textbf{Fear Anxiety \& Dread}} \\
\cmidrule(r){1-2}
  Acute Panic                      & panicked, terrified, alarmed \\
  Agitated Anxiety                 & anxious, stressed, apprehensive, worried, uneasy, agitated, distressed \\
  Visceral Fear                    & horrified, fearful \\
  Threats \& Premonitions          & alarming, alarmist, apocalyptic, ominous, wary, foreboding, cautionary, threatening \\
\addlinespace[1em]
\multicolumn{2}{@{}l@{}}{\textbf{Sorrow Grief \& Regret}} \\
\cmidrule(r){1-2}
  Deep Anguish \& Grief            & sorrowful, mournful, grieving, heartbroken, anguished, devastated \\
  Sadness \& Disappointment        & dismayed, dissatisfied, sad, disappointed \\
  Guilt \& Remorse                 & rueful, contrite, remorseful, ashamed, regretful \\
  Nostalgic Sadness                & wistful, lamenting \\
\addlinespace[1em]
\multicolumn{2}{@{}l@{}}{\textbf{Despair Exhaustion \& Cynicism}} \\
\cmidrule(r){1-2}
  Loss Of Morale                   & discouraged, disheartened, demoralized \\
  Profound Despair                 & desperate, dejected, despairing, despondent \\
  Resignation \& Helplessness      & resigned, helpless, fatalistic \\
  Harsh Gloom                      & bleak, grim \\
  Burnout \& Fatigue               & weary, exhausted \\
  Cynical Disillusionment          & cynical, disillusioned, jaded, pessimistic, hardened \\
  Unbothered Boredom               & bored \\
\addlinespace[1em]
\multicolumn{2}{@{}l@{}}{\textbf{Detachment Restraint \& Defiance}} \\
\cmidrule(r){1-2}
  Coldness \& Callousness          & callous, unsympathetic \\
  Reluctant Concession             & begrudging, reluctant \\
  Self Protection \& Shame         & embarrassed, defensive, sheepish \\
  Defiance \& Rebellion            & rebellious, defiant \\
  Pleading                         & imploring \\
  Gravity \& Silence               & solemn, serious \\
\end{xltabular}
\begin{xltabular}{\textwidth}{@{} >{\raggedright\arraybackslash}p{0.32\textwidth} X @{}}
\caption{Hierarchical Tone Taxonomy: Positive Attitude Domain} \label{tab:tone_taxonomy_positive} \\
\toprule
\textbf{L2 Subgroup} & \textbf{Tone Synonyms} \\
\midrule
\endfirsthead
\caption[]{Hierarchical Tone Taxonomy: Positive Attitude Domain (Continued)} \\
\toprule
\textbf{L2 Subgroup} & \textbf{Tone Synonyms} \\
\midrule
\endhead
\bottomrule
\endlastfoot
\multicolumn{2}{@{}l@{}}{\textbf{Joy \& High Energy}} \\
\cmidrule(r){1-2}
  Pure Happiness                   & glad, pleased, happy, cheerful, gleeful, delighted \\
  Intense Euphoria                 & exuberant, euphoric, ecstatic \\
  Vibrant Excitement               & eager, enthusiastic, carefree, excited, lighthearted \\
\addlinespace[1em]
\multicolumn{2}{@{}l@{}}{\textbf{Affection Warmth \& Care}} \\
\cmidrule(r){1-2}
  Tender Devotion                  & affectionate, adoring \\
  Empathy \& Compassion            & altruistic, thoughtful, sympathetic, compassionate, considerate, empathetic \\
  Social Warmth \& Openness        & welcoming, friendly \\
  Helpfulness \& Tolerance         & helpful, protective, tolerant, permissive, supportive \\
\addlinespace[1em]
\multicolumn{2}{@{}l@{}}{\textbf{Praise Admiration \& Awe}} \\
\cmidrule(r){1-2}
  Deep Reverence \& Awe            & amazed, admiring, awed, impressed, awestruck \\
  Verbal Praise \& Tribute         & laudatory, approving, celebratory, respectful, complimentary, congratulatory, deferential \\
  Gratitude \& Value               & appreciative \\
\addlinespace[1em]
\multicolumn{2}{@{}l@{}}{\textbf{Optimism Hope \& Inspiration}} \\
\cmidrule(r){1-2}
  Forward Looking Confidence       & bullish, optimistic, hopeful, positive \\
  Motivation \& Drive              & encouraging, promotional, proactive, exhortative \\
\addlinespace[1em]
\multicolumn{2}{@{}l@{}}{\textbf{Confidence Conviction \& Triumph}} \\
\cmidrule(r){1-2}
  Strong Conviction                & emphatic, confident, assertive \\
  Passion \& Fervor                & patriotic, impassioned, effusive, passionate \\
  Pride \& Vindication             & triumphant, triumphalist, vindicated, proud, boastful \\
\addlinespace[1em]
\multicolumn{2}{@{}l@{}}{\textbf{Curiosity \& Intellectual Drive}} \\
\cmidrule(r){1-2}
  Active Inquiry                   & inquisitive, curious, intrigued \\
\addlinespace[1em]
\multicolumn{2}{@{}l@{}}{\textbf{Reassurance Harmony \& Satisfaction}} \\
\cmidrule(r){1-2}
  Agreement \& Peace               & agreeable, conciliatory, affirmative, constructive \\
  Making Amends                    & apologetic \\
  Relief \& Trust                  & reassuring, reassured, grateful, receptive, understanding, relieved \\
\addlinespace[1em]
\multicolumn{2}{@{}l@{}}{\textbf{Humor Playfulness \& Sincerity}} \\
\cmidrule(r){1-2}
  Amusement \& Wit                 & humorous, playful, amused, witty \\
  Heartfelt Earnestness            & earnest \\
\end{xltabular}
\begin{xltabular}{\textwidth}{@{} >{\raggedright\arraybackslash}p{0.32\textwidth} X @{}}
\caption{Hierarchical Tone Taxonomy: Neutral Attitude Domain} \label{tab:tone_taxonomy_neutral} \\
\toprule
\textbf{L2 Subgroup} & \textbf{Tone Synonyms} \\
\midrule
\endfirsthead
\caption[]{Hierarchical Tone Taxonomy: Neutral Attitude Domain (Continued)} \\
\toprule
\textbf{L2 Subgroup} & \textbf{Tone Synonyms} \\
\midrule
\endhead
\bottomrule
\endlastfoot
\multicolumn{2}{@{}l@{}}{\textbf{Objective \& Analytical}} \\
\cmidrule(r){1-2}
  Pure Fact \& Logic               & factual, objective, logical \\
  Scientific \& Clinical           & clinical, scientific, technical \\
  Scholarly Analysis               & analytical, academic \\
  Observational Neutrality         & neutral, matter-of-fact, observational \\
  Journalistic Reporting           & reportorial, journalistic \\
  Informational Explanation        & informative, explanatory \\
\addlinespace[1em]
\multicolumn{2}{@{}l@{}}{\textbf{Instructive \& Authoritative}} \\
\cmidrule(r){1-2}
  Educational Instruction          & instructive \\
  Advisory Recommendation          & advisory, suggestive \\
  Normative Direction              & prescriptive \\
  Command Authority                & stern, authoritarian, directive \\
  Formal Protocol                  & official, formal, legalistic \\
  Scholarly Rigidity               & pedantic \\
\addlinespace[1em]
\multicolumn{2}{@{}l@{}}{\textbf{Brevity \& Directness}} \\
\cmidrule(r){1-2}
  Concise Brevity                  & concise, succinct \\
  Plain Directness                 & straightforward \\
  Terse Economy                    & terse, laconic \\
  Open Frankness                   & candid \\
\addlinespace[1em]
\multicolumn{2}{@{}l@{}}{\textbf{Prudence Caution \& Strategy}} \\
\cmidrule(r){1-2}
  Practical Utility                & utilitarian, practical, pragmatic, realistic \\
  Cautious Restraint               & restrained, prudent, measured, cautious \\
  Contingent Framing               & conditional \\
  Calculated Planning              & strategic \\
  Tentative Hesitation             & uncertain, tentative \\
  Alert Watchfulness               & vigilant \\
\addlinespace[1em]
\multicolumn{2}{@{}l@{}}{\textbf{Conviction Resolve \& Urgency}} \\
\cmidrule(r){1-2}
  Firm Resolve                     & certain, determined \\
  Pressing Importance              & urgent \\
  Serious Gravity                  & somber \\
  Reality Inducing Gravity         & sobering \\
\addlinespace[1em]
\multicolumn{2}{@{}l@{}}{\textbf{Detachment Stoicism \& Passivity}} \\
\cmidrule(r){1-2}
  Flat Unconcern                   & apathetic, indifferent \\
  Half Heartedness                 & lukewarm \\
  Unbothered Composure             & unsurprised, nonchalant \\
  Enduring Composure               & patient \\
  Emotional Suppression            & stoic \\
  Impersonal Automation            & robotic, mechanical \\
\addlinespace[1em]
\multicolumn{2}{@{}l@{}}{\textbf{Style Manner \& Mindset}} \\
\cmidrule(r){1-2}
  Conversational Informality       & casual, informal, colloquial \\
  Courteous Etiquette              & polite \\
  Playful Evasiveness              & coy \\
  Obscure Enigma                   & cryptic \\
  Artistic Expression              & poetic \\
  Pensive Introspection            & reflective \\
  Active Querying                  & inquiring \\
  Perplexed Confusion              & puzzled \\
\addlinespace[1em]
\multicolumn{2}{@{}l@{}}{\textbf{Composure Concern \& Foresight}} \\
\cmidrule(r){1-2}
  Attentive Concern                & concerned \\
  Soothing Calm                    & calming \\
  Future Projection                & prophetic \\
\end{xltabular}
\begin{xltabular}{\textwidth}{@{} >{\raggedright\arraybackslash}p{0.32\textwidth} X @{}}
\caption{Hierarchical Tone Taxonomy: Mixed Attitude Domain} \label{tab:tone_taxonomy_mixed} \\
\toprule
\textbf{L2 Subgroup} & \textbf{Tone Synonyms} \\
\midrule
\endfirsthead
\caption[]{Hierarchical Tone Taxonomy: Mixed Attitude Domain (Continued)} \\
\toprule
\textbf{L2 Subgroup} & \textbf{Tone Synonyms} \\
\midrule
\endhead
\bottomrule
\endlastfoot
\multicolumn{2}{@{}l@{}}{\textbf{Firmness Authority \& Conviction}} \\
\cmidrule(r){1-2}
  Unyielding Resolve               & adamant, unapologetic, resolute \\
  Commanding Authority             & authoritative, strict, commanding, imperative \\
  Assertive Insistence             & insistent, forceful, dogmatic \\
\addlinespace[1em]
\multicolumn{2}{@{}l@{}}{\textbf{Uncertainty Conflict \& Hesitation}} \\
\cmidrule(r){1-2}
  Internal Contradiction           & conflicted, ambivalent \\
  Cautious Reserve                 & guarded, hesitant \\
  Hypothetical Conjecture          & speculative \\
  Perplexed Bewildered             & perplexed \\
\addlinespace[1em]
\multicolumn{2}{@{}l@{}}{\textbf{Directness Candor \& Tact}} \\
\cmidrule(r){1-2}
  Unfiltered Candor                & direct, blunt, frank \\
  Strategic Tact                   & diplomatic \\
  Explanatory Precision            & clarifying, nuanced \\
\addlinespace[1em]
\multicolumn{2}{@{}l@{}}{\textbf{Persuasion Rhetoric \& Vision}} \\
\cmidrule(r){1-2}
  Inspirational Drive              & persuasive, visionary, motivational \\
  Ideological Zeal                 & radical \\
  Stylistic Argumentation          & rhetorical \\
\addlinespace[1em]
\multicolumn{2}{@{}l@{}}{\textbf{Surprise Emotion \& Drama}} \\
\cmidrule(r){1-2}
  Sudden Wonder                    & astonished, surprised \\
  Heightened Theater               & melodramatic, dramatic \\
\addlinespace[1em]
\multicolumn{2}{@{}l@{}}{\textbf{Reflection Nostalgia \& Spirituality}} \\
\cmidrule(r){1-2}
  Looking Back                     & anecdotal, nostalgic, retrospective, bittersweet \\
  Looking Forward                  & anticipatory \\
  Transcendent Or Traditional      & superstitious, mystical \\
\addlinespace[1em]
\multicolumn{2}{@{}l@{}}{\textbf{Irreverence Vulgarity \& Playfulness}} \\
\cmidrule(r){1-2}
  Crude Or Coarse                  & crass, vulgar \\
  Childish Or Playful              & immature, punny, joking \\
  Dark Or Self Satisfied           & smug, morbid \\
\addlinespace[1em]
\multicolumn{2}{@{}l@{}}{\textbf{Bias Desire \& Pleading}} \\
\cmidrule(r){1-2}
  Sectional Bias                   & partisan \\
  Possessive Desire                & jealous, envious \\
  Earnest Entreaty                 & pleading \\
\end{xltabular}

\section{Tone Annotation Validation and Inter-Annotator Agreement}
\label{sec:appendix_iaa}

To validate the quality of LLM-based tone annotations, three expert annotators independently evaluated model-generated tone labels across a representative sample of 207 subreddits (one reaction per community). 

\paragraph{Evaluation Protocol}
Annotators evaluated tone suitability on a 4-point rating scale:
\begin{itemize}
    \item \textbf{1.0 (Exact/High Alignment):} Tone captures the exact pragmatic strategy and illocutionary intent of the reaction.
    \item \textbf{0.0 (Acceptable Alignment):} Tone is plausible and contextually fitting, despite minor stylistic variance.
    \item \textbf{-0.5 (Moderate Discrepancy):} Tone is partially misleading or misses key pragmatic nuances.
    \item \textbf{-1.0 (Unacceptable / Wrong):} Tone fundamentally mischaracterizes the speaker's intent.
\end{itemize}
For binary validation reported in the main text, ratings of $\{1.0, 0.0\}$ were categorized as \textit{Accepted}, while $\{-0.5, -1.0\}$ were categorized as \textit{Rejected}.

\paragraph{Label Distributions and Agreement Rates}
\begin{table}[htbp]
  \centering
  \small
  \begin{tabular}{l ccc}
    \toprule
    \textbf{Rating Level} & \textbf{Annotator 1} & \textbf{Annotator 2} & \textbf{Annotator 3} \\
    \midrule
    Exact / High Alignment & 171 & 181 & 187 \\
    Acceptable Alignment & 28 & 17 & 10 \\
    Moderate Discrepancy & 6 & 7 & 7 \\
    Unacceptable / Wrong & 2 & 2 & 3 \\
    \midrule
    \textit{Accepted} & 199 & 198 & 197 \\
    \textit{Rejected} & 8 & 9 & 10 \\
    \bottomrule
  \end{tabular}
  \caption{Annotator label distribution across the 4-way scale and binary acceptance totals.}
  \label{tab:appendix_tone_rate_dist}
\end{table}

\begin{table}[htbp]
  \centering
  \small
  \begin{tabular}{l cc}
    \toprule
    \textbf{Annotator Pair / Group} & \textbf{Binary Agreement (\%)} & \textbf{4-Way Agreement (\%)} \\
    \midrule
    Annotator 1 -- Annotator 2 & $94.7$ & $85.0$ \\
    Annotator 2 -- Annotator 3 & $94.7$ & $89.4$ \\
    Annotator 1 -- Annotator 3 & $95.2$ & $88.4$ \\
    \midrule
    Unanimous (All Three) & $92.3\%$ & $82.6\%$ \\
    \bottomrule
  \end{tabular}
  \caption{Percent agreements of binary and 4-way evaluation setups.}
  \label{tab:appendix_percent_agree}
\end{table}
Table~\ref{tab:appendix_tone_rate_dist} presents the label frequencies across annotators. Table~\ref{tab:appendix_percent_agree} compares pairwise and unanimous percent agreement between the binary and 4-way evaluation schemes.

\paragraph{Prevalence Paradox and Inter-Rater Reliability}
Despite high percent agreement, traditional metrics yield modest scores under binary classification (Fleiss' $\kappa = 0.380$, Krippendorff's $\alpha = 0.381$). This is a classic example of the \textit{high agreement / low Kappa paradox}: under extreme class imbalance ($96.6\%$ positive acceptance), the expected chance agreement in Kappa/Alpha formulas approaches $1.0$, artificially deflating the score.

To correctly account for prevalence, we report Gwet's chance-corrected coefficients~\citep{gwet2008computing}:
\begin{itemize}
    \item \textbf{Binary Setup:} Gwet's $\text{AC}_1 = 0.944$ ($95\%\text{ CI } [0.915, 0.972]$), confirming exceptional inter-rater stability for binary acceptance.
    \item \textbf{4-Way Setup:} Gwet's $\text{AC}_2$ (ordinal weights) $= 0.963$ ($95\%\text{ CI } [0.947, 0.979]$), demonstrating robust consensus even when penalizing fine-grained rating distances.
\end{itemize}

\section{Prompt Templates}
\label{sec:appendix_prompt}

This appendix provides the exact prompt templates used for the generative tasks in our evaluation framework for both tone labeling and simulations. Variables injected at inference time are denoted by curly brackets (e.g., \texttt{\{subreddit\}}).


\subsection{Tone labeling}
\label{app:prompt:tone-labeling-prompts}
\begin{lstlisting}[style=promptstyle]
Hi and welcome! Thank you for taking on this conversational tone labeling task. Here are the instructions:

1. You will be given a text in English. Your task is to identify its production-oriented conversational tone - that is, the author's way of writing/speaking, not necessarily how others might perceive it.

2. You will also receive contextual information to assist your understanding of the given text, including:
   - subreddit: the Reddit community where the text appeared
   - month: the month in 2020 when the text was posted  
   - headline: the related news headline the text is likely responding to
   - excerpt: the first 500 characters of the related news content

3. Take as much time as you need to consider the tone. Your response must be exactly one lowercase English adjective (no other formats will be accepted).

4. If the tone is unclear, do not consult external information. Provide your best guess using the same response format (exactly one lowercase English adjective).

Thank you for carefully following these instructions.

Here is the contextual information for your reference:

subreddit: /r/{subreddit}
month: 2020/{month}
headline: "{news_title}"
excerpt: "{news_content}..."

And here is your text to label the tone of:

"{comment}"
\end{lstlisting}

\subsection{Reaction simulation — community-blind}
\label{app:prompt:comm-blind-simulation-prompt}
\begin{lstlisting}[style=promptstyle]
Hi and welcome! Thank you for participating in this news reaction simulation task. Please follow the instructions below:

1. You will be asked to simulate a Reddit-style comment reacting to a piece of news. The objective is to generate a short, natural-sounding comment that plausibly reflects how some Reddit user might have reacted.

2. You will be provided with the following news information:
    - headline: the headline of the news article being reacted to
    - excerpt: the first 500 characters of the news article

3. Your entire response must consist only of the simulated Reddit comment. Do not include explanations, disclaimers, instructions, or any text outside the comment itself.

4. Do not consult or rely on any external information beyond what is provided. Base your comment solely on the given inputs and your understanding of how a person (reddit user) might respond. If uncertain, provide your best attempt.

Thank you for carefully following these instructions.

Here is the contextual information to simulate a reaction for:

headline: "{news_title}"
excerpt: "{news_content}..."

Your response:
\end{lstlisting}

\subsection{Reaction simulation — community-informed}
\label{app:prompt:comm-informed-simulation-prompt}
\begin{lstlisting}[style=promptstyle]
Hi and welcome! Thank you for participating in this online community news reaction simulation task. Please follow the instructions below:

1. You will be asked to simulate a Reddit-style comment reacting to a piece of news, written as if you were a member of a specified community (subreddit). The objective is to generate a short, natural-sounding comment that plausibly reflects how someone from that community would have reacted.

2. You will be provided with the following contextual information:
    - subreddit: the name of the Reddit community whose reaction you are simulating
    - headline: the headline of the news article being reacted to
    - excerpt: the first 500 characters of the news article

3. Your entire response must consist only of the simulated Reddit comment. Do not include explanations, disclaimers, instructions, or any text outside the comment itself.

4. Do not consult or rely on any external information beyond what is provided. Base your comment solely on the given inputs and your understanding of how the specified community might plausibly respond. If uncertain, provide your best attempt.

Thank you for carefully following these instructions.

Here is the contextual information to simulate a reaction for:

subreddit: /r/{subreddit}
headline: "{news_title}"
excerpt: "{news_content}..."

Your response:
\end{lstlisting}

\subsection{Reaction simulation — community-informed + CoT}
\label{app:prompt:reaction-comm-aware-cot-instructions}
\begin{lstlisting}[style=promptstyle]
Hi and welcome! Thank you for participating in this online community news reaction simulation task. Please follow the instructions below:

1. You will be asked to simulate a Reddit-style comment reacting to a piece of news from 2020, written as if you were a member of a specified community (subreddit). You will be provided with the following contextual information:
    - subreddit: the name of the Reddit community whose reaction you are simulating
    - subreddit description: the self-description of the subreddit as shown on Reddit
    - headline: the headline of the news article being reacted to
    - excerpt: the first 500 characters of the news article

2. The objective is to generate a short, natural-sounding comment that plausibly reflects how someone from that community would have reacted at the time.

3. Before generating your comment, please do a step-by-step analysis of the news and its context, considering the following aspects:
   - The subreddit and its self-description: Consider the community's focus, values, and typical discourse. How might these factors influence the reaction to the news?
   - The headline and excerpt: Analyze the content of the news article, including its subject matter, style, and any potential biases. How might these elements influence the reaction from the community?
   - Anything else you find important to deriving the reaction, such as the presence of specific keywords, phrases, or references that may evoke a particular attitude or perspective.

4. Return only valid JSON. Your response must be parseable by Python's json.loads() function and must exactly follow this structure:

{
  "reasoning": {
    "subreddit_analysis": "",
    "news_content_analysis": "",
    "additional_factors": ""
  },
  "simulated_comment": ""
}

Do not include Markdown, code fences, explanations, or any text outside the JSON object. Use double quotes for all keys and string values, escaping nested double quotes and special characters with backslashes as required by the JSON specification. Use an empty string for a reasoning field only if no relevant analysis is available. Keep each reasoning field concise while including the key considerations that informed the simulated comment.

5. Do not consult any external material. Base your design solely on the given inputs and your understanding of how the specified community might plausibly respond. If uncertain, provide your best attempt.

Thank you for carefully following these instructions.

Here is the contextual information to simulate a reaction for:

subreddit: /r/{subreddit}
subreddit description: "{subreddit_description}"
headline: "{news_title}"
excerpt: "{news_content}..."
\end{lstlisting}

\end{document}